\documentclass[runningheads]{llncs}

\usepackage[T1]{fontenc}
\def\doi#1{\href{https://doi.org/\detokenize{#1}}{\url{https://doi.org/\detokenize{#1}}}}
\usepackage{graphicx}
\usepackage{listings}
\usepackage{enumitem}
\usepackage[numbers]{natbib}
\usepackage{url} 
\usepackage{hyperref}

\hypersetup{
  colorlinks   = true, 
  urlcolor     = blue, 
  linkcolor    = blue, 
  citecolor   = blue 
}
\usepackage[dvipsnames]{xcolor}
\usepackage{xspace}
\usepackage{soul,color}
\soulregister\cite7
\soulregister\ref7
\soulregister\pageref7
\setstcolor{red}
\setul{}{.2ex}

 \usepackage{verbatim}
 \usepackage{multicol}
 \usepackage[utf8]{inputenc}
 \usepackage{amsmath} \allowdisplaybreaks
 \usepackage{cleveref}
 \usepackage{amssymb}
 \usepackage{array}
 \usepackage{booktabs}
\usepackage{caption}
\usepackage[numbers]{natbib}
\usepackage{mathtools}
\usepackage{csvsimple}
\usepackage{caption}
\usepackage{subcaption}
\usepackage{multirow}
\usepackage{siunitx}

\usepackage{glossaries} 
\glsdisablehyper 
\newacronym{mh}{MH}{Metaheuristic}
\newacronym{ls}{LS}{Local Search}
\newacronym{ts}{TS}{Tabu Search}
\newacronym{sa}{SA}{Simulated Annealing}
\newacronym{lahc}{LAHC}{Late Acceptance Hill Climbing}
\newacronym{hls}{HLS}{Hybrid Local Search}
\newacronym{cp}{CP}{Constraint Programming}
\newacronym{ilp}{ILP}{Integer Linear Programming}
\newacronym{lns}{LNS}{Large Neighborhood Search}
\newacronym{uc1}{UC-1}{Use Case 1}
\newacronym{uc2}{UC-2}{Use Case 2}

\newacronym{stn}{STN}{Search Trajectory Network}
\newacronym{shap}{SHAP}{SHapley Additive exPlanations}
\newacronym{cd}{CD}{Critical Difference}

\newacronym{swcb}{SCB}{Swap Consecutive Batches}
\newacronym{ib}{IB}{Insert Batch}
\newacronym{mvj}{MJEB}{Move Job to Existing Batch}
\newacronym{mjnb}{MJNB}{Move Job to New Batch}
\newacronym{swb}{SB}{Swap Batches}
\newacronym{mj}{MJ}{Move Jobs}
\newacronym{ibo}{IBO}{Invert Batches Order}

\newacronym{mssp}{MSSP}{Medical Student Scheduling Problem}
\newacronym{pfsp}{PFSP}{Permutation Flowshop Scheduling Problem}
\newacronym{tsp}{TSP}{Traveling Salesperson Problem}
\newacronym{gcp}{\emph{k}-GCP}{\emph{k}-Graph Coloring Problem}
\newacronym{osp}{OSP}{Oven Scheduling Problem}

\newacronym{rbm}{RB1M}{Random Batches from  One Machine}
\newacronym{rb}{RB}{Random Batches}
\newacronym{rba}{RBA}{Random Batches with the Same Attribute}
\newacronym{opl}{OPLRUN}{CP Optimizer model}

\usepackage{enumitem}
\newlist{ResearchQuestions}{enumerate}{1}
\setlist[ResearchQuestions]{label=\textbf{RQ\arabic*},left=0pt, widest=10}

\begin{document}
\title{Theoretical Lower Bounds for the\\ Oven Scheduling Problem}
\titlerunning{Theoretical Lower Bounds for the OSP}

\author{
Francesca Da Ros\inst{1}\orcidID{0000-0001-7026-4165}
\and
Marie-Louise Lackner\inst{2}\orcidID{0000-0002-9916-9011}\and
Nysret Musliu\inst{2}\orcidID{0000-0002-3992-8637}}
\institute{
DMIF, University of Udine, Italy
\email{francesca.daros@uniud.it}\and
Christian Doppler Laboratory for Artificial Intelligence and Optimization for Planning and Scheduling, Institute for Logic and Computation, TU Wien, Austria
\email{\{marie-louise.lackner, nysret.musliu\}@tuwien.ac.at}}

\authorrunning{F.\ Da Ros et al.}

\maketitle              

\begin{abstract}
The Oven Scheduling Problem (OSP) is an NP-hard real-world parallel batch scheduling problem arising in the semiconductor industry. 
The objective of the problem is to schedule a set of jobs on ovens while minimizing several factors, namely total oven runtime, job tardiness, and setup costs. 
At the same time, it must adhere to various constraints such as oven eligibility and availability, job release dates, setup times between batches, and oven capacity limitations.
The key to obtaining efficient schedules is to process compatible jobs simultaneously in batches. 
In this paper, we develop theoretical, problem-specific lower bounds for the OSP that can be computed very quickly. 
We thoroughly examine these lower bounds, evaluating their quality and exploring their integration into existing solution methods.
Specifically, we investigate their contribution to exact methods and a metaheuristic local search approach using simulated annealing.
Moreover, these problem-specific lower bounds enable us to assess the solution quality for large instances for which exact methods often fail to provide tight lower bounds.

\end{abstract}

\section{Introduction}
\label{sec:intro}

The semiconductor manufacturing sector has been identified as one of the most energy-intensive industries \cite{wang-2023-semiconductor}, particularly in the context of hardening electronic components in specialized heat treatment ovens. 
To mitigate energy consumption, one strategy involves grouping and processing compatible jobs together in batches to optimize resource utilization.
Such scheduling tasks that aim to increase efficiency by processing multiple jobs simultaneously in batches are known as 
batch scheduling problems.

Over the last three decades, the scientific community has extensively investigated batch scheduling problems, as witnessed by the surveys by~\citet{mathirajan2006literature, fowler_survey_2022}.
A multitude of problem variants, in the single or parallel machine setting,  and each with distinct constraints and objectives imposed by different industries \cite{tang-2020-manu,zaho-2020-steel} have been studied. 
One such formulation, the \gls{osp}, was recently introduced by \citet{lackner-2023-exact} and is particularly pertinent to semiconductor manufacturing.
The goal of this problem is to efficiently schedule jobs on multiple ovens, aiming to minimize total oven runtime, job tardiness, and setup costs simultaneously. 
In order to reach these goals, 
 compatible jobs are grouped and processed together in batches. 
Schedules must adhere to various constraints, including oven eligibility and availability, job release dates, setup times between batches, oven capacity limitations, and compatibility of job processing times.

The \gls{osp} was initially addressed using exact methods as well as a heuristic construction method: \citet{lackner-2023-exact} proposed two different modeling approaches, encompassing \gls{cp} and \gls{ilp} model formulations.
The exact approaches successfully identified optimal solutions for 38 out of 80 benchmark instances.
However, for larger instances, optimal solutions were rarely obtained within a time-bound of one hour.
In a later extended abstract, a metaheuristic local search approach based on \gls{sa} was suggested by~\citet{lackner-2022-sa}.
This approach showed promising results, as optimal solutions could often be reached quickly and non-optimal solutions were improved for numerous instances. 

In practical settings, it is most often desirable to obtain solutions of sufficiently good, albeit not necessarily optimal, quality within a short time frame. 
However, assessing the solution quality becomes challenging in the absence of a baseline, i.e., when exact methods are not employed 
or do not deliver tight enough lower bounds on the objective value.
Providing problem-specific, efficiently computable lower bounds on the optimal solution cost can thus be very helpful in assessing the quality of a solution. 
Moreover, lower bounds can aid existing solution approaches and increase their performance:
in exact methods, they can be used to bound the range of variables, and in (meta-)heuristic search methods, they can be included in stopping criteria.
Theoretical, problem-specific lower bounds have been developed for batch scheduling problems in the literature.
\citet{damodaran2012simulated} proposed a \gls{sa} approach in a parallel batch setting and presented a procedure for calculating lower bounds on the makespan. Additionally, lower bounds on the makespan and total completion time have been addressed by \citet{Koh_Koo_Kim_Hur_2005}. The maximum lateness has been tackled by \citet{li_heuristics_2019,Kedad-Sidhoum_Solis_Sourd_2008}. While these lower bounds have been proposed for different batching problems, not all features of the \gls{osp} have been considered previously.   

\begin{figure}[ht]
    \centering
    \includegraphics[width=0.95\textwidth]{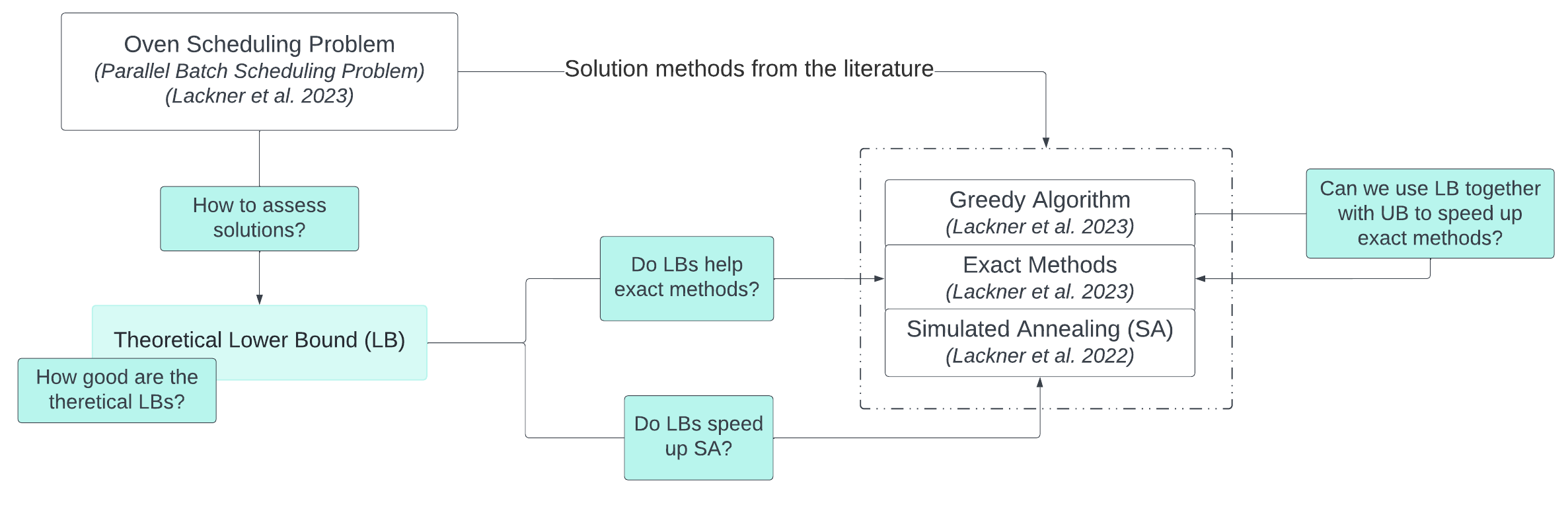}
    \caption{Overview of the goals targeted by this work.}
    \label{fig:visual-abstract}
\end{figure}

In this paper, we present an approach to computing lower bounds for the objectives of the \gls{osp}, making use of the imposed machine eligibility and processing time constraints.
The goals of this paper are visualized in \Cref{fig:visual-abstract}.
Our primary 
 contributions are as follows:
\begin{itemize}[topsep=0pt]
    \item We introduce a procedure to compute theoretical lower bounds for the \gls{osp}, more specifically for the number of batches and the total oven runtime.  These lower bound results can be adapted to tackle related (parallel) batch scheduling problems. Our approach differs from the existing literature as it considers machine eligibility and compatibility of processing times.
    
    \item We conduct a comprehensive evaluation of the tightness of our calculated lower bounds on a benchmark set consisting of 120 instances with up to 500 jobs. This evaluation encompasses the overall cost function and its individual components.     
    We differentiate between instances where an optimal solution is available and those where it is not. Notably, for larger instances with 50 jobs or more, our calculated lower bounds provide a small gap w.r.t. the optimal solution value and very often outperform the lower bounds generated by commercial solvers (when the optimal solution value is not known).
    
    \item We integrate the derived lower bounds into state-of-the-art solution approaches and demonstrate that they
    can aid with solving the \gls{osp}. 
    Our experiments explore to what extent exact methods benefit from being provided with the calculated lower bounds.
    Furthermore, we investigate whether lower bounds can speed up 
    \gls{ls} algorithms, such as \gls{sa}. 
    Using a 1\% gap between the \gls{sa} solution and the calculated lower bound as a stopping criterion, many of the benchmark instances can be solved very fast (50 of the 120 benchmark instances are solved in roughly 15 seconds on average).
    
    \item To encourage future contributions and enhance the 
    replicability of results, we provide a software toolbox that enables the generation of instances and the calculation of lower bounds.
\end{itemize}

\noindent
The remainder of this paper is structured as follows.
\Cref{sec:oven-scheduling-problem} introduces the \gls{osp}.
\Cref{sec:lower-bounds} elaborates on the theoretical calculations of lower bounds for the \gls{osp}.
\Cref{sec:including-lower-bounds} displays proposals on how to integrate lower bounds in the solution methods.
\Cref{sec:experimental-evaluation} details our experimental evaluation.
Eventually, \Cref{sec:conclusion} draws some conclusions and suggests future research directions.

\section{The Oven Scheduling Problem}
\label{sec:oven-scheduling-problem}

The \gls{osp} aims to group compatible jobs into batches and devise an optimal schedule for these batches across a set of ovens. 
We report an abridged description of the problem and forward the interested reader to the rigorous mathematical formulation proposed in the original paper~\cite{lackner-2023-exact}.

An instance of the \gls{osp} consists of a set $\mathcal{M}=\{1,\dots,k\}$ of ovens (also referred to as machines) as well as a set $\mathcal{A}=\{1, \dots, a\}$ of possible attributes (also known as job families in the literature). Each machine $m \in \mathcal{M}$ is associated with a maximal processing capacity $c_m$ and an initial state $ia_m \in \mathcal{A}$. 
Each oven presents a set of availability intervals $[as(m,i), ae(m,i)]$, where $as(m,i)$ ($ae(m,i)$) indicates the start (end) of the $i$-th interval. 

A set $\mathcal{J} = \{1, \dots, n\}$ of jobs is given. Each job $j \in \mathcal{J}$ is described by an attribute $a_j \in \mathcal{A}$, a size $s_j \in \mathbb{N}$,  
an earliest start time (or release date) $et_j \in \mathbb{N}$, 
and a latest end time (or due date) $lt_j \in \mathbb{N}$. 
The processing of a job is constrained by its minimal and maximal processing times ($mint_j$ and $maxt_j \in \mathbb{N}$, respectively). 
Additionally, jobs have eligibility constraints, limiting their assignment to specific machines (indicated with the set $\mathcal{E}_j \subseteq \mathcal{M}$). 

Setup times and costs are incurred between consecutive batches on the same machine and depend upon the attributes of the batches (attributes of jobs in the batch). They are indicated with two $(a \times a)$-matrices of setup times $st=(st(a_i, a_j))_{1 \leq a_i, a_j \leq a}$ and of setup costs $sc=(sc(a_i, a_j))_{1 \leq a_i, a_j \leq a}$ are given to denote the setup times (costs) incurred between a batch with attribute $a_i$ and a subsequent one with attribute $a_j$.

The \gls{osp} aims to establish a feasible assignment of jobs to ovens, grouping them into batches, and to determine the schedule of batches on the ovens.
A feasible batch construction and schedule must respect the following rules:

\begin{itemize}[topsep=0pt]
    \item \textbf{Attribute homogeneity}: Jobs in the same batch must share the attribute.

    \item \textbf{Release date}: A batch cannot start processing until the release date of the latest-released job assigned to it.

    \item \textbf{Processing time}: The processing time of a batch must be longer than or equal to the minimal processing time and shorter than or equal to the maximal processing time of any job in the batch. 
    Jobs in the same batch start and finish processing at the same time and job-preemption is not allowed.

    \item \textbf{Setup time}: Batches on the same machine may not overlap, and setup times between consecutive batches need to be respected.

    \item \textbf{Machine eligibility}: Jobs can only be assigned to one of their eligible machines. 

    \item \textbf{Machine availability}: For every batch,  the entire processing time and the preceding setup time must be scheduled within a single availability interval. 

    \item \textbf{Machine capacity}: The size of each batch cannot exceed the capacity of the machine it is assigned to.
\end{itemize}
 
The objective of the \gls{osp} is threefold: to minimize the cumulative batch processing time ($p$), the number of tardy jobs ($t$), and the cumulative setup costs ($sc$). 
Given a solution to the \gls{osp}, the three objective components are formally defined as follows: 

\begin{align*}
p &= \sum_{m \in \mathcal{M}} \sum_{b \in \mathcal{B}_m}  P_{m,b},\quad &
t &= \left \vert \left\{ j \in \mathcal{J} 
: C_j > lt_j\right\} \right \vert & \text{and }
sc &=  \sum_{m \in \mathcal{M}} \sum_{b \in \mathcal{B}_m} sc_{m,b}
\end{align*}
where $\mathcal{B}_m$ is the set of batches of machine $m \in \mathcal{M}$ that composes the solution. 
The processing time of batch $b \in \mathcal{B}_m$ on machine $m \in \mathcal{M}$ is indicated with $P_{m,b}$.
The total tardiness is calculated as the number of jobs $j \in \mathcal{J}$ for which the completion time $C_j$ is greater than their due date $lt_j$ in the given solution.
The setup cost of batch $b$ on machine $m$ is denoted by $sc_{m,b}$.
Each component is then normalized 
and aggregated in a weighted sum
to account for different real-world scenarios. 
The weights used throughout this paper are set as follows: $w_p=4$, $w_t=100$, and $w_{sc}=1$ (these are also normalized by their sum, see Use Case 1 by \citet{lackner-2023-exact}). 
To illustrate the problem, \Cref{sec:example-osp-instance} reports an example instance for the \gls{osp}.

\subsection{Solution methods for the \gls{osp}}
\label{sec:solution-methods-for-osp}

In the literature, the \gls{osp} has been solved with a construction heuristic \cite{lackner-2023-exact}, exact methods \cite{lackner-2023-exact}, and a \gls{sa} algorithm \cite{lackner-2022-sa} which we very briefly describe here. 

The construction heuristic introduced to solve the \gls{osp}~\cite{lackner-2023-exact} is a dispatching rule that prioritizes
jobs based on their release dates and then on their due dates.
The algorithm starts at time 0. 
At each time step, it compiles the list of currently available machines and currently released jobs that have not yet been scheduled.  
The algorithm then selects the job with the earliest due date from this pool and greedily assigns it to one of the eligible machines. 
Once a job is scheduled, other available jobs are included in the same batch, provided that the job's attribute, processing time, and the machine's capacity allow it. If no job can be scheduled, the time is incremented by one, and the process is repeated.
This heuristic has been used to warm-start the exact methods with some of the solvers~\cite{lackner-2023-exact} and as an initial solution for the \gls{sa} approach~\cite{lackner-2022-sa}.


Two exact modeling approaches which were formulated as \gls{cp} and \gls{ilp} models were proposed by~\citet{lackner-2023-exact}.
The first approach
is based on batch positions: each job is assigned to one of the possible batches, which are uniquely characterized by their machine and the batch position on this machine.
The constraints are formulated on the level of batches and an optimal schedule of the batches needs to be found.
The second 
uses a unique representative job for each batch and seeks an optimal schedule for these jobs.  
These two modeling approaches are 
implemented both in the high-level solver-independent modeling language
MiniZinc \cite{nethercote_minizinc_2007} and using interval variables in the Optimization
Programming Language (OPL)~\cite{hentenryck2002constraint} used by CP Optimizer.
Moreover, different state-of-the-art solvers, search strategies, and a warm-start approach leveraging the construction heuristic were employed. Ultimately, the best results were achieved with \verb|CP Optimizer| and the OPL-model using representative jobs as well as with \verb|Gurobi| and the MiniZinc-model with batch positions. 
In what follows, we will refer to these two solution methods as ``cpopt'' and ``mzn-gurobi'' (as well as ``cpopt-WS'' and ``mzn-gurobi-WS'' for the variants with warmstart).

A \gls{sa} algorithm for the \gls{osp} was proposed by~\citet{lackner-2022-sa}.
In this algorithm, a solution to the \gls{osp} 
is represented by the assignments of jobs to ovens and by the processing order of the jobs on their respective machines. 
The schedule of the batches on the ovens is then deterministically constructed from this representation.
The initial solution is retrieved from the construction heuristic previously presented.
The algorithm relies on four neighborhood-moves: 
the \gls{swcb} move, which swaps consecutive batches on the same machine;
the \gls{ib} move, which inserts a given batch in a new position on the same machine;
the \gls{mvj} move, which inserts a job $j$ in an existing batch;
the \gls{mjnb} move, which inserts a job in a newly created batch. 
In the original work by~\citet{lackner-2022-sa}, \gls{sa} was proposed with a preliminary manual tuning, whereas we fine-tuned its parameters for this work.

\section{Lower bounds on the optimal solution cost}
\label{sec:lower-bounds}

In this section, we describe a procedure to calculate lower bounds on the optimal solution cost for a given instance of the \gls{osp}. 
Our main focus lies in bounding the number of batches required in any feasible solution. 
At the same time, we derive bounds on the cumulative batch processing time.
These lower bounds serve as a basis for deriving lower bounds on the cumulative setup costs. 
Finally, we provide a brief discussion on the number of tardy jobs.

\subsection{Minimum number of batches required and minimal cumulative batch processing time}
\label{sec:lower-bounds-batch-number}

Since jobs can only be combined in a batch if they share the same attribute, bounds on the number of batches required are calculated independently for all attributes. 
For a given attribute $r \in \mathcal{A}$, 
we denote by $b_r$ the number of batches in a feasible solution and by
$p_r$ the minimal cumulative processing time of batches.

\subsubsection{Bound based on machine capacities and job sizes.}

Due to the capacity constraints of machines, a simple bound on the number of batches required is
\begin{equation}
 b_r \geq \left\lceil  \frac{\sum_{j \in \mathcal{J}: a_j = r}s_j}{\max_{m \in \mathcal{M}}\{c_m\}} \right \rceil,  
 \label{eqn:lower-bound-machine-cap}
\end{equation}
as stated by~\citet{Koh_Koo_Kim_Hur_2005}.
This corresponds to the minimal number of batches required if we assume that jobs can be split into smaller jobs of unit size and that all jobs can be scheduled on the machine with the largest machine capacity.

This bound can be tightened by distinguishing between ``large'' and ``small'' jobs (in a similar fashion as~\citet{damodaran2012simulated, Li_Chen_Du_Tan_2013, li_heuristics_2019}).
Large jobs are those jobs that are so large that they cannot accommodate any other jobs in the same batch and thus need to be processed in a batch of their own.
All other jobs are referred to as small jobs.
For a given attribute $r$, the sets of large jobs $J_r^l$ and small jobs $J_r^s$ with attribute $r$ are thus defined as follows:
\begin{align*}  
J_r^l &= \left\lbrace j \in \mathcal{J}: a_j = r,  s_j + s_i >\max_{m \in \mathcal{E}_j}(c_m) \quad \forall i \in \mathcal{J} \text{ with } i \neq j \text{ and } a_i =r\right\rbrace, \\
J_r^s & = \left\lbrace j \in \mathcal{J}: a_j = r\right \rbrace \setminus J_r^l
\end{align*}
Instead of the bound in equation~\eqref{eqn:lower-bound-machine-cap}, we thus have the tighter bound:
\begin{equation}
b_r \geq \vert J_r^l \vert + \left\lceil  \frac{\sum_{j \in J_r^s}s_j}{\max_{m \in \mathcal{M}}\{c_m\}} \right \rceil.
     \label{eqn:better-lower-bound-machine-cap}
\end{equation}

In the following, we refine these bounds from the literature by considering machine eligibility and compatibility of processing times.

\subsubsection{Refinement of the bound for small jobs based on machine eligibility.}

Considering the small jobs of attribute $r \in \mathcal{A}$, we further distinguish them between those that can be processed on several machines and those with a single eligible machine.  
Given a machine $i \in \mathcal{M}$,  we use the 
following notation:
\[
b_{r,i} = \frac{
\sum_
{j \in J_r^s:  \mathcal{E}_j=\{i\}}
s_j}
{c_i}, \text{ and }
cap_i = ( \lceil b_{r,i} \rceil -b_{r,i} ) \cdot c_i
\]
i.e., $\lceil b_{r,i} \rceil$ is the minimal number of batches with small jobs that need to be processed on machine $i$ 
and $cap_i$ is the total remaining 
capacity in these batches.

To schedule the small jobs of attribute $r$, we proceed as follows: 
\begin{itemize}[topsep=0pt]
    \item All small jobs that need to be processed on a specific machine are scheduled on this machine.
    \item The remaining small jobs are used to fill up the previously created batches.
    \item If there are still jobs left, we assume that they can be split into unit-size jobs and can be scheduled on the machine with maximal capacity, creating $b_r^*$ additional batches. 
\end{itemize}
The bound in equation~\eqref{eqn:better-lower-bound-machine-cap} can then be tightened as follows:
\begin{align}
b_r & \geq  b_r^E =\vert J_r^l \vert + \sum_{i \in \mathcal{M}} \lceil b_{r,i} \rceil + \underbrace{\left\lceil \frac{\max{(0,\sum_{j \in J_r^s: \vert \mathcal{E}_j \vert > 1}s_j - \sum_{i \in \mathcal{M}}{cap_i})}}{\max_{m \in \mathcal{M}}\{c_m\}} \right\rceil}_{=b_r^*}
    \label{eqn:even-better-lower-bound-small-jobs-elig-machines}
\end{align}

In order to calculate a lower bound on the cumulative batch processing time $p_r$ of these batches, 
note that all large jobs are processed in batches of their own which run for their respective minimal processing times. Thus 
\begin{equation}
    p_r = \sum_{j \in J_r^l}mint_j + p_r^E,
    \label{eqn:lower-bound-runtime-el-mach}
\end{equation}
where $p_r^E$ denotes the minimal cumulative processing time of batches consisting of small jobs with attribute $r$. A bound for $p_r^E$ can be calculated as follows:
\begin{itemize}[topsep=0pt]
    \item For every machine $i$ with $b_{r,i}>0$, create the collection of minimal processing times of small jobs that need to be processed on $i$; create the sum of the $\lceil b_{r,i} \rceil$ smallest elements from this collection.
    \item From the collection of minimal processing times of small jobs that can be processed on multiple machines, create the sum of the $b_r^*$ smallest elements.
    \item Among all small jobs, pick the one with the largest minimal processing time. The batch containing this job will necessarily have this job's minimal processing time. In the previous two sums, one can thus replace the overall largest processing time with this value.
\end{itemize}
\vspace{-0.2cm}

\subsubsection{Alternative refinement of the bound for small jobs based on compatible job processing times.}
\label{sec:bounds-comp-processing-times}

Two jobs $i$ and $j$ with respective minimal and maximal processing times $mint_i, mint_j$ and $maxt_i, maxt_j$
may only be combined in a batch if the intervals of their processing times have a non-empty intersection:
\begin{equation}
[mint_i, maxt_i] \cap [mint_j, maxt_j] \neq \emptyset.
\label{eqn:comp-proc-time}
\end{equation}
This compatibility relation between jobs can be represented with the help of a \textit{compatibility graph} $G=(V,E)$, where $V$ is the set of all jobs $\mathcal{I}$ and $(i,j) \in E$ if and only if the jobs $i$ and $j$ have compatible processing times. 
In this graph, a batch forms a (not necessarily maximal) clique.
The problem of solving an OSP instance with unit-sized jobs and a single machine with capacity $c$ is thus equivalent to covering the nodes of the compatibility graph with the smallest number of cliques with size no larger than $c$.

This problem is NP-complete for arbitrary graphs, but solvable in polynomial time for interval graphs. A simple greedy algorithm is provided  by~\citet{finke2008batch} and referred to as the algorithm GAC (greedy algorithm with compatibility).
By adapting the order in which jobs are processed by the GAC algorithm, we obtain an algorithm that minimizes both the number of batches and the cumulative batch processing time.
We call this algorithm GAC+. 

\textit{Algorithm GAC+}: 
Consider the jobs in non-increasing order $j_1, j_2, \ldots, j_n$  of their minimal processing times $mint_j$, breaking ties arbitrarily.
Construct one batch per iteration until all jobs have been placed into batches.
In iteration $i$, open a new batch $B_i$ and label it with the first job $j^*$ that has not yet been placed in a batch. Starting with $j^*=[mint_{j^*},maxt_{j^*}]$, place into $B_i$ the first $c$ not yet scheduled jobs $j$ for which $mint_{j^*} \in [mint_{j},maxt_{j}]$. 

For a set $\mathcal{J}$ of jobs with arbitrary job sizes, let $GACb(\mathcal{J},c)$ denote the number of batches returned by the GAC+ algorithm when replacing every job $j \in \mathcal{J}$ with $s_j$ identical copies of unit size jobs. Similarly, let $GACp(\mathcal{J},c)$ denote the minimal processing time returned by the GAC+ algorithm for this instance. With this notation, we obtain  the following bounds:
\begin{align}
b_r & \geq \vert J_r^l \vert +  b_r^C, &\text{ with } b_r^C = GACb(J_r^s,\max_{m \in \mathcal{M}}\{c_m\}), 
    \label{eqn:even-better-lower-bound-small-jobs-proc-times} \\
p_r & \geq \sum_{j \in J_r^l}mint_j + p_r^C, &\text{ with } p_r^C = GACp(J_r^s,\max_{m \in \mathcal{M}}\{c_m\}).
\label{eqn:even-better-lower-bound-runtime-small-jobs-proc-times}
\end{align}
For a formal statement and proof of this result, see Section~\ref{sec:appendix-GAC+} of the appendix.

\subsubsection{Overall bound on the number of batches and the minimal cumulative processing time.}

Combining the previously established bounds, we obtain:
\[
b  \geq \sum_{r=1}^a (\vert J_r^l \vert + \max(b_r^E, b_r^C))  \text{ and }
p  \geq \sum_{r=1}^a(\sum_{j \in J_r^l}mint_j + \max(p_r^E, p_r^C)),
\]
where $b_r^E$ is defined in equation~\eqref{eqn:even-better-lower-bound-small-jobs-elig-machines} and $b_r^C$ in equation~\eqref{eqn:even-better-lower-bound-small-jobs-proc-times}, the procedure to calculate $p_r^E$ is described right after equation~\eqref{eqn:lower-bound-runtime-el-mach} and $p_r^C$ is defined in equation~\ref{eqn:even-better-lower-bound-runtime-small-jobs-proc-times}.

\subsection{Bounds on the other components of the objective function}
\label{sec:lower-bounds-obj-components}

\subsubsection{Setup costs.}

If we assume that the setup costs \textit{before} batches of a given attribute are always minimal, we obtain the following bound on the setup costs:
\begin{equation}
sc \geq \sum_{r=1}^a b_r \cdot \min_{s \in \{1, \ldots, a \}}\{ sc(s,r) \}.
\label{eqn:lower-bound-setup-costs-before-batch}
\end{equation}

A similar bound can be derived assuming that the setup costs \textit{after} batches are always minimal. 
For this case, we include initial setup costs for all machines to which batches are scheduled and ignore the last batch on every machine.
Since a prior it is not known which machines are used in a schedule, we create the list \texttt{setup\_costs} as follows. 
For every attribute $r$, we add $b_r$ copies of $\min_{s \in \{1, \ldots, a \}}\{ sc(r,s) \}$ to \texttt{setup\_costs}. 
Moreover, for every machine $m$, we add the element $\min_{s \in \{1, \ldots, a \}}\{ sc(ia_m,s) \}$ to \texttt{setup\_costs}.
The list is then sorted in non-decreasing order and the sum of the first $b$ elements is taken: 
\begin{equation}
sc \geq \sum_{i = 1}^b \texttt{setup\_costs}(i).
\label{eqn:lower-bound-setup-costs-after-batch}
\end{equation}
Altogether, we have the following lower bound on the setup costs
\begin{equation}
    sc \geq \max
    \left( \sum_{r=1}^a b_r \cdot \min_{s \in \{1, \ldots, a \}}\{ sc(s,r) \}, 
    \sum_{i = 1}^b \texttt{setup\_costs}(i)
    \right).
    \label{eqn:lower-bound-setup-costs}
\end{equation}

Note that it is impossible to obtain a lower bound on the setup costs 
by arranging the minimum number of batches per attribute (as calculated previously) in an order that minimizes the cumulative setup costs.
Indeed, if the matrix of setup costs does not fulfill the triangle inequality, it can be advantageous to introduce additional batches if the sole objective is to reduce setup costs. 

\subsubsection{Number of tardy jobs.}

Regarding the number of tardy jobs, direct inference from the instance itself may be limited. 
However, we can obtain a lower bound on the number of tardy jobs by independently scheduling each job in a batch on its own on the first available machine and computing the completion time. 
Any job finishing after its latest end date is necessarily tardy in every solution.

\section{Including lower bounds in solution methods}
\label{sec:including-lower-bounds}

A recommended practice 
to build efficient exact models is to tightly restrict and bound the domain of variables (as suggested, for instance, by the MiniZinc guide on efficient modeling practices\footnote{see \url{https://www.minizinc.org/doc-2.5.5/en/efficient.html}}). 

By employing tighter variable bounds, algorithmic efficiency can be significantly enhanced, facilitating faster convergence to optimal solutions or the identification of unfeasible regions.
When solving the \gls{osp} with one of the exact methods, the lower bounds derived in Section~\ref{sec:lower-bounds} can be calculated in a preprocessing step and can then be provided to the model as part of the input data. The range of the variables corresponding to the individual objective components as well as the variable for the aggregated objective function can thus be bounded from below. Moreover, the aggregated objective value of the solution delivered by the construction heuristic can be used to bound the range of the objective function from above.

Problem-specific lower bounds can also have practical applications in metaheuristic algorithms, e.g., in \gls{sa}.
Lower bounds can be used to guide the search, e.g., as part of the termination criterion.
This strategy allows for early interruption of the process, sparing computational resources while still achieving satisfactory solution quality 

\section{Experimental evaluation}
\label{sec:experimental-evaluation}

In this experimental evaluation, we aim to analyze 
the quality of the theoretically derived lower bounds and their practical usefulness in helping to solve the \gls{osp}.

\subsection{Benchmark instances}
\label{sec:instances}

We consider the 80 benchmark instances by \citet{lackner-2022-instances}, which differ per number of jobs  (10, 25, 50, or 100), number of machines (2 or 5), and number of attributes (2 or 5).

Moreover, we consider 40 new instances featuring a larger number of jobs (250 or 500) to reflect real-world scenarios better. This new set is generated using the specifications of the random instance generator provided by \citet{lackner-2021-cp}.
The instances can be retrieved from the public 
public GitHub repository 
 \url{https://github.com/iolab-uniud/osp-ls/}.

For tuning purposes (i.e., when using \gls{sa}), we generate 25 additional instances with similar characteristics as the initial benchmark set. 

\subsection{Experimental setup}
\label{sec:experimental-setup}

We consider the following methods for the \gls{osp}:
\begin{itemize}[topsep=0pt]
    \item \textbf{Problem-specific lower bounds}  (presented in \Cref{sec:lower-bounds}):
    For the instances we consider, the bounds are calculated in 2.9 seconds on average (with a standard deviation of 6.9 s). 
    \item \textbf{Construction heuristic} (proposed by \citet{lackner-2023-exact}, see \Cref{sec:solution-methods-for-osp}): 
    Since the solution is deterministically constructed, there is no need to execute the algorithm more than once.  For the instance we consider, the solutions are retrieved in 0.2 seconds on average (with a standard deviation of 0.4 s).
    \item \textbf{Best performing exact methods} (proposed by \citet{lackner-2023-exact}, see \Cref{sec:solution-methods-for-osp}): 
    We refer to the methods as ``cpopt'' (interval variable model with representative jobs solved with \verb|CP Optimizer|) and ``mzn-gurobi'' (MiniZinc-model with batch positions solved with \verb|Gurobi|), as well as ``cpopt-WS'' and ``mzn-gurobi-WS'' for the variants with warm-start.
    Each method is run with a timeout of 1 hour per instance. 
    \item \textbf{Local search approach with \gls{sa}}(proposed by \citet{lackner-2022-sa}, see \Cref{sec:solution-methods-for-osp}):
    The algorithm is tuned using automated parameter tuning with \verb|irace|~\cite{lopex-ibanez-2016-irace}.  
To account for the stochastic components of \gls{sa}, we execute the algorithm 10 times per instance with a timeout of 6 minutes.
Every 2 seconds we record the overall cost and the single objective components of the best solution encountered so far.
\end{itemize}
Details regarding the implementation, the tuned parameters of the \gls{sa} and the hardware can be found in Section~\ref{sec:details-exp-setup}
 of the appendix.

\subsection{Lower bounds quality}
\label{sec:lower-bound-quality}

Our objective is to assess the tightness of the calculated lower bounds. 
We examine the bound on the overall cost ($obj$) as well as the bounds on its three components individually ($t$, $p$, and $sc$). 
For the smaller benchmark instances with up to 100 jobs and the aggregated objection function, we refer to the best results per instance obtained by~\citet{lackner-2023-exact} with their proposed exact methods.
For the larger benchmark instances with 250 or 500 jobs, we rerun the best-performing exact methods (``mzn-gurobi'' and ``mzn-gurobi-WS'' as well as ``cpopt'' and ``cpopt-WS'') and retrieve the best result per instance.
Moreover, we run the exact models with the task of optimizing just one of the three components for the entire benchmark set.
In our analysis of the lower bounds, we differentiate between those instances and objectives where an optimal solution cost is known and those where we do not know the optimum.  

\begin{figure}[ht]
    \centering
    \includegraphics[width=0.9\textwidth]{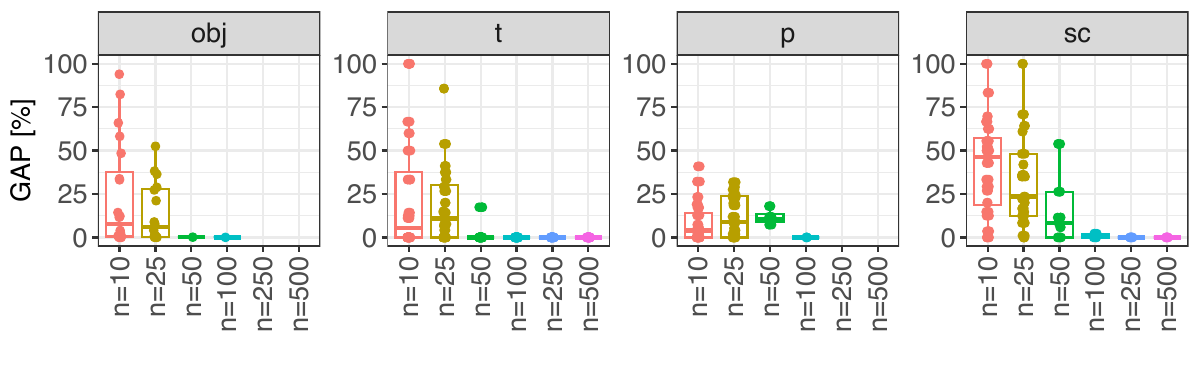}
    \caption{Gap[\%] between the known optimum and the calculated lower bounds.}
    \label{fig:opt-vs-bound}
\end{figure}

For those instances and objectives where the optimum solution is known, given an instance $i$, we compute the relative gap$(i)$ between the calculated lower bound $b(i)$ and the optimal cost $s(i)$; specifically
$\text{gap}(i) = 100 \cdot (s(i)-b(i))/s(i)$. 
Results show the general tendency that the larger the instances, the smaller the gap (see \Cref{fig:opt-vs-bound}). 
Concerning the individual components, we observe that most room for improvement is left for the simple bounds for $sc$ and $t$. Nonetheless, the gap for $sc$ is less than $25\%$ for more than half of the instances and the gap for $t$ is less than $10\%$ for $74\%$ of the instances. 
For the cumulative processing times, 
the gap is less than $25\%$ for $88\%$ of instances and less than $10\%$ for $61\%$. The results are promising, as they give reason to hope that the bounds are relatively tight for instances where the optimum is not known as well.

Whenever the optimal solution value is not known, we compare the problem-specific lower bounds with the lower bounds retrieved by \verb|CP Optimizer| and \verb|Gurobi|  (specifically, ``cpopt'', ``cpopt-WS'', ``mzn-gurobi'', and ``mzn-gurobi-WS'') 
and retrieve the best, i.e., largest, lower bound found per instance. 
For each objective, we count how often the calculated lower bounds are better, worse, or equal to the best dual bounds found by the exact methods, see~\Cref{tab:lb-calculated-vs-lb-solver}.
The results show that both for the overall cost and its components, the calculated problem-specific lower bounds are better than those provided by any of the exact methods in the majority of the instances. The dominance of the problem-specific lower bounds is particularly clear for the larger instances with 100 jobs or more. 
Interestingly, this observation holds even for the objective components ``setup costs'' (problem-specific bounds are better or equally good in 2/3 of the instances) and ``number of tardy jobs'' (better or equally good results in 94 \% of the instances) for which the calculated bounds are very simple.

Moreover, we investigated the gap between the calculated lower bounds and the upper bounds provided by the construction heuristic (see \Cref{sec:solution-methods-for-osp}). 
For a total of 57 instances, this gap is less than 10\% (see \Cref{sec:upper-bound} for details).

\begin{table}[ht]
    \centering
    \scriptsize
    \caption{Comparison of the quality of calculated problem-specific  (``calc.'') and best solver lower bounds (``solv.''). We consider only those instances for which no optimal solution is known. The label ``calc.''  refers to the number of instances where the calculated bounds are better, ``solv.'' to those where the solver bounds are better and ``equ.'' to those where the bounds are equal.}
    \label{tab:lb-calculated-vs-lb-solver}
    \begin{tabular}{
    l @{\hspace{2mm}}
    S[table-format=1.] @{\hspace{1mm}}
    S[table-format=1.] @{\hspace{1mm}}
    S[table-format=1.]
    c @{\hspace{2mm}}
    S[table-format=1.] @{\hspace{1mm}}
    S[table-format=1.] @{\hspace{1mm}}
    S[table-format=1.]
    c @{\hspace{2mm}}
    S[table-format=1.] @{\hspace{1mm}}
    S[table-format=1.] @{\hspace{1mm}}
    S[table-format=1.]
    c @{\hspace{2mm}}
    S[table-format=1.] @{\hspace{1mm}}
    S[table-format=1.] @{\hspace{1mm}}
    S[table-format=1.]
}
    \toprule
     & \multicolumn{3}{c}{$obj$} & & \multicolumn{3}{c}{$t$} & & \multicolumn{3}{c}{$p$} & & \multicolumn{3}{c}{$sc$} \\
    \cmidrule(lr){2-4} \cmidrule(lr){6-8} \cmidrule(lr){10-12} \cmidrule(lr){14-16}
     & {calc.} & {solv.} & {equ.} && {calc.} & {solv.} & {equ.} && {calc.} & {solv.} & {equ.} && {calc.} & {solv.} & {equ.}\\
     $n$ & {\#} & {\#} & {\#} && {\#} & {\#} & {\#}  && {\#} & {\#} & {\#}  && {\#} & {\#} & {\#} \\
    \midrule
     25  & 0  & 4  & 0 && 0  & 0 & 0  && 1  & 0 & 0 && 1  & 0 & 0 \\
     50  & 7  & 12 & 0 && 0  & 6 & 2  && 10 & 6 & 0 && 4  & 7 & 2 \\
     100 & 16 & 3  & 0 && 1  & 6 & 0  && 18 & 1 & 0 && 15 & 1 & 2 \\
     250 & 20 & 0  & 0 && 10 & 1 & 1  && 20 & 0 & 0 && 18 & 0 & 0 \\
     500 & 20 & 0  & 0 && 10 & 0 & 2  && 20 & 0 & 0 && 18 & 0 & 0 \\
     \midrule
     all & 63  & 19  & 0 && 21  & 13 & 5  && 69  & 7 & 0 && 56  & 8 & 4 \\
    \bottomrule
\end{tabular}
\end{table}

\subsection{Measuring solution quality}
\label{sec:measuring-solution-quality}

\begin{figure}[ht]
    \centering    \includegraphics[width=0.9\textwidth]{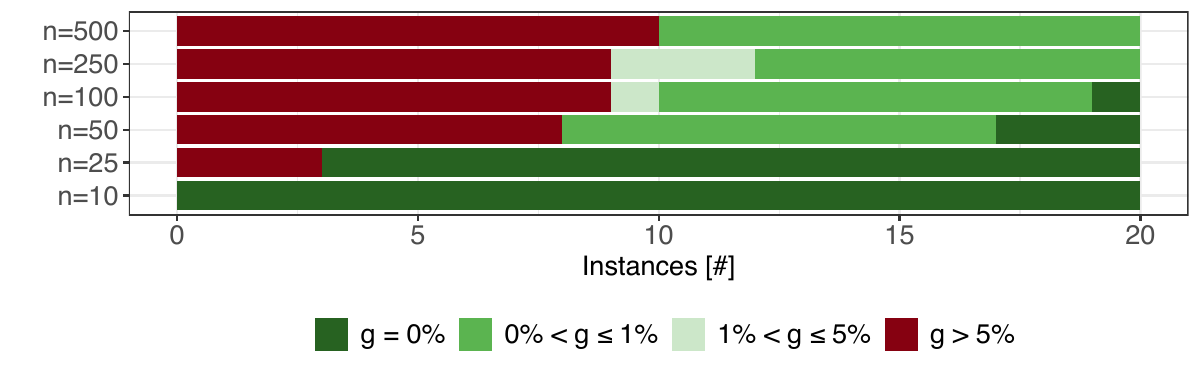}
    \caption{Gap[\%] between the best solution found and the best lower bounds.}
    \label{fig:rq-3counter-per-instance-size}
\end{figure}

In this section, we use the lower bounds to assess the solution quality and benchmark the best-known solutions for the \gls{osp} with the best lower bounds. 
On the one hand, we consider the best solution found for each instance by the methods described in \Cref{sec:solution-methods-for-osp}. 
On the other hand, we consider the best lower bound per instance among the calculated bounds and the ones retrieved by the exact methods.
Then we calculate the relative gap between the best solution and the best lower bound per instance. 
Results are shown in \Cref{fig:rq-3counter-per-instance-size}.
Almost all small instances with 25 or 50 jobs could be solved optimally. For larger instances, the solution methods find very good solutions (with a relative $\text{gap}[\%] \leq 1\%$) for roughly half the instances. For most of the remaining instances, the gap is larger than 5\%, showing that there is still room for improvement--both in terms of the solution quality and in terms of the lower bound quality.

\subsection{Application of lower bounds}

\subsubsection{Exact methods}
\label{sec:experiments-em}

We aim to understand whether using the calculated problem-specific lower bounds allows the exact methods to improve their results. 
As described in Section~\ref{sec:including-lower-bounds}, we perform experiments where the objective function and its components are bounded from below by the calculated lower bounds.
Moreover, we perform experiments additionally supplying the solvers with the upper bound on the objective obtained from the greedy construction heuristic.

\Cref{tab:stats_all_exact_methods_withorwithoutbounds} presents results categorized by methods and types of bounds included; it displays: 
the number of instances for which the optimal solution, when known, was reached (``optimal'');
the number of instances for which a feasible solution was found (``solved''),
the number of instances for which the method could prove optimality (``proven opt'');
the number of instances for which the best solution was found (``best'');
the number of instances for which the best lower bound could be found (``best lower bound'');
the average run time (``avg rt'') and its standard deviation (``std rt'') in seconds. 
Note that for the number of best solutions found and of best lower bounds found, the comparison is made among a single solution method, i.e., comparing results obtained when no non-trivial bounds are provided, when lower bound and when lower and upper bounds are provided.
The statistics regarding runtime are calculated for the subset of instances for which the respective solution method could prove optimality when it was not provided with bounds  (meaning that instances for which the time-out was reached are not included).
The majority of solution methods, namely ``mzn-gurobi'', ``cpopt'' and ``cpopt-WS'', demonstrate greatly improved performance and solution quality when lower bounds are incorporated. 
For ``mzn-gurobi-WS''\footnote{The warm-start data provided to Gurobi only contains values for a subset of the decision variables. The solver
thus needs to complete the partial solution and, for ``mzn-gurobi-WS', fails to do so for many large instances.}, the contribution of the bounds is less clear: fewer instances are solved (optimally), but better solutions and better lower bounds can be found.
The inclusion of upper bounds is not always advantageous for the exact methods, meaning that the solvers were not capable of finding a solution that was at least as good as the greedy solution within a time limit of 1 hour.
For all analyzed solution methods, the presence of bounds facilitates the discovery of improved lower bounds by the commercial solvers, thus contributing to closing the optimality gap.

\begin{table}[ht]
    \centering
    \scriptsize
    \caption{Comparison of the results obtained with exact methods 
    with and without the inclusion of bounds. Best results per solution method and performance parameter are highlighted in bold font. Numbers in brackets indicate the improvement obtained by supplying the respective solution methods with bounds.}
    \label{tab:stats_all_exact_methods_withorwithoutbounds}
    \begin{tabular}{
    l@{\hspace{2mm}}
    l@{\hspace{2mm}}
    l@{\hspace{2mm}}
    l@{\hspace{2mm}}
    l@{\hspace{2mm}}
    l@{\hspace{2mm}}
    l@{\hspace{2mm}}
    r@{\hspace{2mm}}
    r}
\toprule
   solution  &  bounds incl.\  &optimal &  solved & proven opt  &  best & best LB  &  avg rt &  std rt \\
   method&  in model & \# & \# & \# & \# & \#  & (in s) & (in s)  \\
\midrule
        mzn-gurobi &    none &   40 &       64 &          31 &       51 &                41 &   429.5 &   860.6 \\
 & LB & \textbf{41} (+1) & \textbf{78} (+14) &     \textbf{36} (+5) & 62 (+11) &          \textbf{62} (+21) &   \textbf{189.9} &   387.6 \\
   & LB + UB & 40 (+0) &  73 (+9) &     35 (+4) & \textbf{64} (+13) &          55 (+14) &   235.8 &   542.0 \\
\midrule
       mzn-gurobi &   none &    \textbf{41} &       \textbf{89} &          \textbf{34} &       57 &                40 &   764.3 &  1217.9 \\
 -WS&LB & 40 (-1) & 87 (-2) &     \textbf{34} (+0) & \textbf{68} (+11) &          65 (+25) &   505.4 &  1069.8 \\
 & LB + UB &  \textbf{41} (+0) & 84 (-5) &     \textbf{34} (+0) &  65 (+8) &          \textbf{70} (+30) &   \textbf{493.9} &  1083.2 \\
\midrule
        cpopt & none&       39 &       \textbf{114} &          28 &        73 &                28 &    18.4 &    34.1 \\
 &LB &  \textbf{40} (+1) &  \textbf{114} (+0) &     \textbf{33} (+5) &   \textbf{79} (+6) &         \textbf{118} (+90) &    \textbf{17.8} &    46.4 \\
 & LB + UB  &  39 (+0) & 85 (-29) &     \textbf{33} (+5) & 57 (-16) &         110 (+82) &    17.9 &    43.1 \\
\midrule
        cpopt-WS &   none &    38 &     \textbf{120} &          28 &       70 &                28 &    17.6 &    30.0 \\
 & LB &  \textbf{40} (+2) & \textbf{120} (+0) &     \textbf{33} (+5) & 81 (+11) &         \textbf{118} (+90) &    \textbf{15.9} &    31.7 \\
  & LB + UB &  \textbf{40} (+2) & \textbf{120} (+0) &     \textbf{33} (+5) & \textbf{83} (+13) &         117 (+89) &    19.9 &    42.9 \\
\bottomrule
\end{tabular}
\end{table}

\Cref{tab:stats_best_exact_vs_best_exact_with_bounds} offers a comprehensive comparison of overall best results. 
The inclusion of bounds enabled the methods to deliver three new optimality proofs and to find 23 better solutions.
Additionally, the computational time reduces when bounds are utilized compared to when they are not.

\begin{table}[ht]
    \centering
    \scriptsize
    \caption{Overall comparison of the best results per instance achieved with exact methods without the inclusion of bounds and with the inclusion of bounds.}    
    \label{tab:stats_best_exact_vs_best_exact_with_bounds}
\begin{tabular}{
    l@{\hspace{2mm}}
    l@{\hspace{2mm}}
    l@{\hspace{2mm}}
    l@{\hspace{2mm}}
    l@{\hspace{2mm}}
    l@{\hspace{2mm}}
    r@{\hspace{2mm}}
    r}
\toprule
    bounds & optimal & solved & proven opt & best & {best lower bound} &  {avg rt} &  {std rt} \\
    included & \# & \# & \# & \# & {\#} &  {(in s)} & {(in s)} \\
\midrule
no &      41 &      120 &          38 &       76 &                42 &   486.8 &  1075.6 \\
   yes &  41 (+0) & 120 (+0) &     41 (+3) & 99 (+23) &         116 (+74) &   107.6 &  256.9 \\
\bottomrule
\end{tabular}
\end{table}

\subsubsection{Local search}
\label{sec:experiments-ls}

\begin{figure}[ht]
        \centering
        \includegraphics[width=0.9\textwidth]{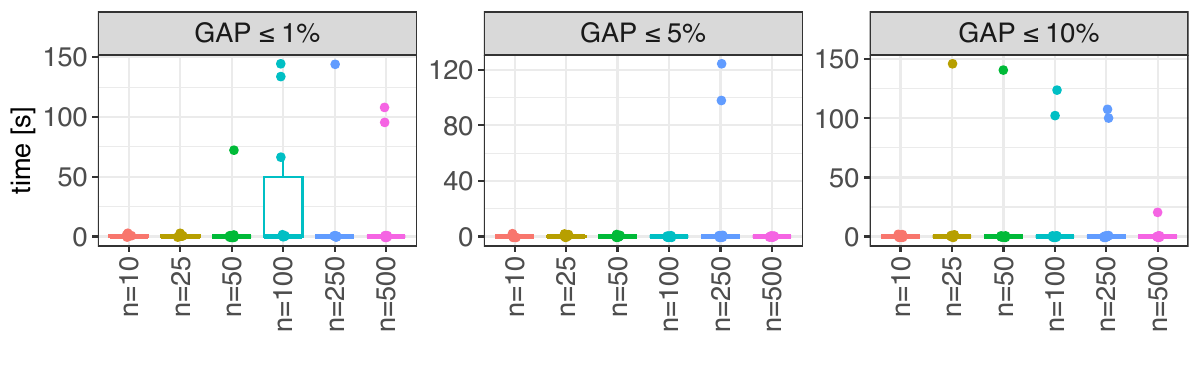}
        \caption{Minimum time required by \gls{sa} to reach a given gap[\%] w.r.t. $obj$.} 
        \label{fig:optimality_gap_sa_time}
\end{figure}

Lower bounds provide a means to assess whether it is feasible to halt the search before reaching the termination criterion -- in our case, the timeout.
We aim to discern under which circumstances this is viable and how much time is necessary.
Considering the overall cost, for 50 out of 120 instances, the gap[\%] is lower than 1\% (average time required 15.52 $\pm$ 39.85 s);
for 60, the gap[\%] is lower than 5\% (average time required 3.86 $\pm$ 20.21 s), and for 67, it is lower than 10\% (average time required 11.13 $\pm$ 34.92 s). 
This means that for roughly half of the benchmark instances, the search could be terminated early, delivering a solution of good quality.
It is worth pointing out that this is merit also of a demonstrably good initial solution (see \Cref{sec:upper-bound}).
\Cref{fig:optimality_gap_sa_time} reports the distribution of minimum time required by \gls{sa} to achieve such results. 

\section{Conclusion}
\label{sec:conclusion}

In this study, we introduced a procedure for calculating theoretical lower bounds for the \gls{osp} which can be calculated within a couple of seconds even for large instances. 
The experimental evaluation demonstrated their quality and practical utility when incorporated into exact methods or \gls{ls} approaches.
Our bounds can help to find better solutions, to deliver more optimality proofs, and to find high-quality solutions in a shorter time.

Notably, some of the bounds we developed are relatively simple, in particular those concerning job tardiness.
This suggests that there is potential for further enhancements by refining these lower bounds with more sophisticated methods.
Therefore, future extensions will focus on improving the presented bounds. 
Additionally, we aim to explore adaptive local search techniques, wherein neighborhood probabilities dynamically adjust based on the proximity to the lower bounds. 
Moreover, investigating alternative use cases, such as employing different weight sets on the objective function, may offer valuable insights.
\medskip

\begin{small}
\noindent
\textbf{Replicability.} 
The software toolbox 
can be retrieved at the public GitHub repository 
\url{https://github.com/marielouiselackner/OvenSchedulingCLI}, and the new benchmark instances are available at \url{https://github.com/iolab-uniud/osp-ls/}.

\medskip
\noindent
\textbf{Acknowledgments.}
The financial support by the Austrian Federal Ministry of Labour and Economy, the National Foundation for Research, Technology and Development and the Christian Doppler Research
Association, and SPECIES is gratefully acknowledged.

\end{small}

\bibliographystyle{splncs04nat}
\bibliography{ms}

\begin{thebibliography}{19}
\providecommand{\natexlab}[1]{#1}
\providecommand{\url}[1]{\texttt{#1}}
\providecommand{\urlprefix}{URL }
\expandafter\ifx\csname urlstyle\endcsname\relax
  \providecommand{\doi}[1]{doi:\discretionary{}{}{}#1}\else
  \providecommand{\doi}{doi:\discretionary{}{}{}\begingroup \urlstyle{rm}\Url}\fi

\bibitem[{Damodaran and V{\'e}lez-Gallego(2012)}]{damodaran2012simulated}
Damodaran, P., V{\'e}lez-Gallego, M.C.: A simulated annealing algorithm to minimize makespan of parallel batch processing machines with unequal job ready times. Expert systems with Applications \textbf{39}(1), 1451--1458 (2012)

\bibitem[{Di~Gaspero and Schaerf(2003)}]{digaspero-2003-easylocal}
Di~Gaspero, L., Schaerf, A.: \textsc{EasyLocal++}: An object-oriented framework for flexible design of local search algorithms. Software --- Practice \& Experience \textbf{33}(8), 733--765 (July 2003)

\bibitem[{Finke et~al.(2008)Finke, Jost, Queyranne, and Seb{\H{o}}}]{finke2008batch}
Finke, G., Jost, V., Queyranne, M., Seb{\H{o}}, A.: Batch processing with interval graph compatibilities between tasks. Discrete Applied Mathematics \textbf{156}(5), 556--568 (2008)

\bibitem[{Fowler and M{\"o}nch(2022)}]{fowler_survey_2022}
Fowler, J.W., M{\"o}nch, L.: A survey of scheduling with parallel batch (p-batch) processing. European Journal of Operational Research \textbf{298}(1), 1--24 (Apr 2022)

\bibitem[{Hentenryck(2002)}]{hentenryck2002constraint}
Hentenryck, P.V.: Constraint and integer programming in {OPL}. INFORMS Journal on Computing \textbf{14}(4), 345--372 (2002)

\bibitem[{Kedad-Sidhoum et~al.(2008)Kedad-Sidhoum, Solis, and Sourd}]{Kedad-Sidhoum_Solis_Sourd_2008}
Kedad-Sidhoum, S., Solis, Y.R., Sourd, F.: Lower bounds for the earliness--tardiness scheduling problem on parallel machines with distinct due dates. European Journal of Operational Research \textbf{189}(3), 1305--1316 (2008)

\bibitem[{Koh et~al.(2005)Koh, Koo, Kim, and Hur}]{Koh_Koo_Kim_Hur_2005}
Koh, S.G., Koo, P.H., Kim, D.C., Hur, W.S.: Scheduling a single batch processing machine with arbitrary job sizes and incompatible job families. International Journal of Production Economics \textbf{98}(1), 81--96 (2005)

\bibitem[{Lackner et~al.(2021)Lackner, Mrkvicka, Musliu, Walkiewicz, and Winter}]{lackner-2021-cp}
Lackner, M.L., Mrkvicka, C., Musliu, N., Walkiewicz, D., Winter, F.: {Minimizing Cumulative Batch Processing Time for an Industrial Oven Scheduling Problem}. In: Michel, L.D. (ed.) 27th International Conference on Principles and Practice of Constraint Programming (CP 2021), Leibniz International Proceedings in Informatics (LIPIcs), vol. 210, pp. 37:1--37:18, Schloss Dagstuhl -- Leibniz-Zentrum f{\"u}r Informatik, Dagstuhl, Germany (2021)

\bibitem[{Lackner et~al.(2022{\natexlab{a}})Lackner, Mrkvicka, Musliu, Walkiewicz, and Winter}]{lackner-2022-instances}
Lackner, M.L., Mrkvicka, C., Musliu, N., Walkiewicz, D., Winter, F.: {Benchmark instances and models for the Oven Scheduling Problem [Data Set]} (Dec 2022{\natexlab{a}}), \doi{10.5281/zenodo.7456938}

\bibitem[{Lackner et~al.(2023)Lackner, Mrkvicka, Musliu, Walkiewicz, and Winter}]{lackner-2023-exact}
Lackner, M.L., Mrkvicka, C., Musliu, N., Walkiewicz, D., Winter, F.: Exact methods for the oven scheduling problem. Constraints \textbf{28}(2), 320--361 (2023)

\bibitem[{Lackner et~al.(2022{\natexlab{b}})Lackner, Musliu, and Winter}]{lackner-2022-sa}
Lackner, M.L., Musliu, N., Winter, F.: Solving an industrial oven scheduling problem with a simulated annealing approach. In: Proceedings of the 13th International Conference on the Practice and Theory of Automated Timetabling, pp. 115--120 (2022{\natexlab{b}})

\bibitem[{Li et~al.(2013)Li, Chen, Du, and Tan}]{Li_Chen_Du_Tan_2013}
Li, X., Chen, H., Du, B., Tan, Q.: Heuristics to schedule uniform parallel batch processing machines with dynamic job arrivals. International Journal of Computer Integrated Manufacturing \textbf{26}(5), 474--486 (2013)

\bibitem[{Li et~al.(2019)Li, Li, and Huang}]{li_heuristics_2019}
Li, X., Li, Y., Huang, Y.: Heuristics and lower bound for minimizing maximum lateness on a batch processing machine with incompatible job families. Computers \& Operations Research \textbf{106}, 91--101 (Jun 2019)

\bibitem[{L{\'o}pez-Ib{\'a}{\~n}ez et~al.(2016)L{\'o}pez-Ib{\'a}{\~n}ez, Dubois-Lacoste, {P{\'e}rez C{\'a}ceres}, Birattari, and St{\"u}tzle}]{lopex-ibanez-2016-irace}
L{\'o}pez-Ib{\'a}{\~n}ez, M., Dubois-Lacoste, J., {P{\'e}rez C{\'a}ceres}, L., Birattari, M., St{\"u}tzle, T.: The irace package: Iterated racing for automatic algorithm configuration. Operations Research Perspectives \textbf{3}, 43--58 (2016)

\bibitem[{Mathirajan and Sivakumar(2006)}]{mathirajan2006literature}
Mathirajan, M., Sivakumar, A.I.: A literature review, classification and simple meta-analysis on scheduling of batch processors in semiconductor. The International Journal of Advanced Manufacturing Technology \textbf{29}(9-10), 990--1001 (2006)

\bibitem[{Nethercote et~al.(2007)Nethercote, Stuckey, Becket, Brand, Duck, and Tack}]{nethercote_minizinc_2007}
Nethercote, N., Stuckey, P.J., Becket, R., Brand, S., Duck, G.J., Tack, G.: {MiniZinc}: {Towards} a {Standard} {CP} {Modelling} {Language}. In: Bessi{\`e}re, C. (ed.) Principles and {Practice} of {Constraint} {Programming} {\textendash} {CP} 2007, pp. 529--543, Lecture {Notes} in {Computer} {Science}, Springer, Berlin, Heidelberg (2007)

\bibitem[{Tang and Beck(2020)}]{tang-2020-manu}
Tang, T.Y., Beck, J.C.: Cp and hybrid models for two-stage batching and scheduling. In: Hebrard, E., Musliu, N. (eds.) Integration of Constraint Programming, Artificial Intelligence, and Operations Research, pp. 431--446, Springer International Publishing, Cham (2020)

\bibitem[{Wang et~al.(2023)Wang, Huang, Chen, Chen, Cai, and Wu}]{wang-2023-semiconductor}
Wang, Q., Huang, N., Chen, Z., Chen, X., Cai, H., Wu, Y.: Environmental data and facts in the semiconductor manufacturing industry: An unexpected high water and energy consumption situation. Water Cycle \textbf{4}, 47--54 (2023)

\bibitem[{Zhao et~al.(2020)Zhao, Liu, Zhou, Guo, and Qi}]{zaho-2020-steel}
Zhao, Z., Liu, S., Zhou, M., Guo, X., Qi, L.: Decomposition method for new single-machine scheduling problems from steel production systems. IEEE Transactions on Automation Science and Engineering \textbf{17}(3), 1376--1387 (2020)

\end{thebibliography}

\newpage

\appendix
\section{Appendix}

\subsection{Example of an \gls{osp} Instance.}
\label{sec:example-osp-instance}

To better exemplify the problem, let us consider the following randomly created instance consisting of 10 jobs ($n=10$), 2 machines ($k=2$), and 2 attributes ($a=2$). It presents the following characteristics:

\begin{center}
\scriptsize
\begin{minipage}[ht]{0.45\textwidth}
\centering
\begin{tabular}{p{1.3cm}|p{1.5cm}p{1.5cm}}
$m$ & $M_1$ & $M_2$   \\\hline
$c_m$ & 18 & 20 \\\hline
$ia_m$ & 1 & 2 \\\hline
$[as, ae]$ & [21,250]  & [103,259]
\end{tabular}
\end{minipage}
\hfill
\begin{minipage}[ht]{0.45\textwidth}
\centering
\[ st = \left(  \begin{array}{cc}
0 & 0 \\
3 & 8 
\end{array} \right)
\qquad
sc =  \left( \begin{array}{cc}
6 & 8  \\
10 & 10
\end{array} \right)
\]
\end{minipage}

\vspace{0.3cm}

\begin{tabular}{l|cccccccccc}
$j$ & 1 & 2 & 3 & 4 & 5 & 6 & 7 & 8 & 9 & 10 \\\hline
$\mathcal{E}_j$ & $M_1$ & $M_1$ &  & $M_1$ & $M_1$ & & $M_1$ & $M_1$ & &$M_1$\\
& $M_2$ & $M_2$   & $M_2$ & &$M_2$ & $M_2$ & $M_2$& &$M_2$ &$M_2$ \\\hline
$et_j$ & 2 & 3 &8&1&39&41&40&31&27&16\\
$lt_j$  &  16&20&43&24&55&64&56&89&58&27 \\
$mint_j$  & 11&10&19&19&10&19&11&50&19&11\\
$maxt_j$    &  11&50&19&19&50&50&50&50&19&50\\
$s_j$     &  18&16&17&2&6&19&11&11&4&14\\
$a_j$      &  2&2&2&1&2&2&2&2&1&1\\
\end{tabular}
\end{center}

\begin{figure}[ht]
    \centering
    \includegraphics[width=0.87\textwidth]{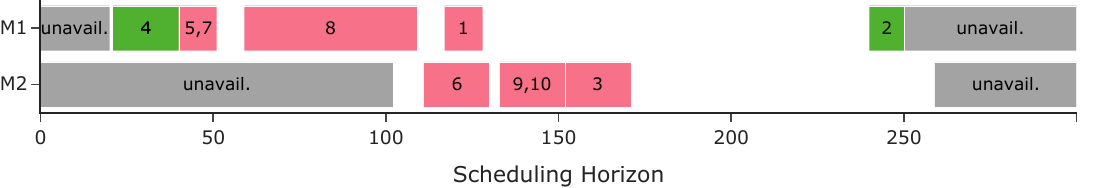}
    \caption{Gantt chart of a solution of the \gls{osp} for an instance with 10 jobs. The label of each bar represents the jobs processed in the batch. Unavailabilities (``unavail.'') are reported in \textcolor{gray}{gray}. Batches with attribute 1 are colored in \textcolor{ForestGreen}{green}, whereas those referring to attribute 2 are colored in \textcolor{magenta}{magenta}. 
    }
    \label{fig:gantt-example}
\end{figure}

\Cref{fig:gantt-example} reports a possible solution to such an instance in the form of a Gantt Chart. 
The running time of the oven is $p = 158$, the number of tardy jobs is $t = 8$, and the setup costs amount to $sc = 72$. 
This solution is optimal when setting the weights in the objective function as $w_p=4$, $w_t=100$, and $w_{sc}=1$.

\subsection{Formal statement and proof of the correctness of the GAC-bounds described in Section~\ref{sec:lower-bounds-batch-number}}
\label{sec:appendix-GAC+}

In the following, we formulate the bounds described in the Section entitled \textit{Alternative refinement of the bound for small jobs based on compatible job processing times} (starting on page \pageref{sec:bounds-comp-processing-times}) more formally and prove the correctness of the algorithm GAC+. 

First, let us recall the compatibility requirement expressed in equation~\eqref{eqn:comp-proc-time}.
Two jobs $i$ and $j$ with respective minimal and maximal processing times $mint_i, mint_j$ and $maxt_i, maxt_j$
may only be combined in a batch if the intervals of their processing times have a non-empty intersection:
\begin{equation}
[mint_i, maxt_i] \cap [mint_j, maxt_j] \neq \emptyset. \notag
\end{equation}

Now let us consider the following special case of the OSP:\begin{quote}
OSP*: Given a set of jobs $\mathcal{I}$ of unit size defined by their minimal and maximal processing times, i.e. $j=[mint_j, maxt_j]$ for all $j \in\mathcal{I}$, and a single machine with capacity $c \in \mathbb{N}$, how many batches do we need at least in order to process all jobs if jobs can only be processed in the same batch if the compatibility condition~\eqref{eqn:comp-proc-time} is fulfilled?\end{quote}
Several variants of this problem have been studied in the literature, e.g.\ by Finke et al.~\cite{finke2008batch}; the variant that we are interested in corresponds to the problem (P2) there.
Solving the problem OSP* will allow us to obtain lower bounds for the OSP: Indeed, as in equation~\eqref{eqn:lower-bound-machine-cap}, we obtain lower bounds on the number of batches required and their processing times if we assume that jobs can be split into smaller jobs of unit size and that all jobs can be scheduled on the machine with largest machine capacity.

As stated in \Cref{sec:lower-bounds}, equation~\eqref{eqn:comp-proc-time} between jobs can be represented with the help of a \textit{compatibility graph} $G=(V,E)$, where $V$ is the set of all jobs $\mathcal{I}$ and $(i,j) \in E$ if and only if the jobs $i$ and $j$ have compatible processing times.
In this graph, a batch forms a (not necessarily maximal) clique.
The problem of solving an OSP instance with unit-sized jobs and a single machine with capacity $c$ is thus equivalent to covering the nodes of the compatibility graph with the smallest number of cliques with size no larger than $c$.

A simple greedy algorithm to solve this problem is provided  by~\citet{finke2008batch} and referred to as the algorithm GAC (greedy algorithm with compatibility).
By adapting the order in which jobs are processed by the GAC algorithm, we obtain an algorithm that minimizes both the number of batches and the cumulative batch processing time.
We call this algorithm GAC+. 

\textit{Algorithm GAC+.}
Consider the set of jobs $\mathcal{I}$ in non-increasing order $j_1, j_2, \ldots, j_n$  of their minimal processing times $mint_j$, breaking ties arbitrarily.
Construct one batch per iteration until all jobs have been placed into batches.
In iteration $i$, open a new batch $B_i$ and label it with the first job $j^*$ that has not yet been placed in a batch. Starting with $j^*=[mint_{j^*},maxt_{j^*}]$, place into $B_i$ the first $c$ not yet scheduled jobs $j$ for which $mint_{j^*} \in [mint_{j},maxt_{j}]$ (or all of them if there are fewer than $c$).

For a set $\mathcal{I}$ of unit size jobs and a maximum batch size $c \in \mathbb{N}$, we denote by $GACb(\mathcal{I},c)$ the number of batches returned by the GAC+ algorithm above.  Similarly, let $GACp(\mathcal{I},c)$ denote the minimal processing time returned by the GAC+ algorithm for this instance.

\begin{theorem}
For any given set of unit size jobs $\mathcal{I}$ and for any given constant $c \in \mathbb{N}$, Algorithm GAC+ solves the problem OSP*, i.e., $GACb(\mathcal{I},c)$ is the minimum number of batches required under the condition that a batch may not contain more than $c$ jobs. Moreover, the cumulative batch processing time $GACp(\mathcal{I},c)$ is minimal.
\label{thm:algo-GAC+}
\end{theorem}

By slight abuse of notation, for a set $\mathcal{J}$ of jobs with arbitrary job sizes, let $GACb(\mathcal{J},c)$ denote the number of batches returned by the GAC+ algorithm when replacing every job $j \in \mathcal{J}$ with $s_j$ identical copies of unit size jobs. Similarly, let $GACp(\mathcal{J},c)$ denote the minimal processing time returned by the GAC+ algorithm for this instance.
With this notation, Theorem~\ref{thm:algo-GAC+} yields the bounds reported in~\eqref{eqn:even-better-lower-bound-runtime-small-jobs-proc-times}.

\begin{proof}[of Theorem~\ref{thm:algo-GAC+}]
We follow the proof of Theorem~4 in~\cite{finke2008batch}, extending it to include the minimization of the cumulative batch processing time and adapting it to our variant of the algorithm.
The proof is by induction over the number of jobs and the induction start with a single job is trivial.

Let us start with a simple observation about the minimum number of batches and the minimal batch processing time.
For this, let $b(\mathcal{I},c)$ denote the minimum number of batches required to schedule all jobs in $\mathcal{I}$ under the condition that a batch may not contain more than $c$ jobs. Similarly, let $p(\mathcal{I},c)$ denote the minimal cumulative batch processing time in any schedule of all jobs in $\mathcal{I}$.
Then these two functions are monotonous in $\mathcal{I}$, i.e.:
\begin{align}
\begin{split}
&b(\mathcal{I},c) \geq b(\mathcal{I} \setminus \{j \},c) \\
\text{and } &p(\mathcal{I},c) \geq p(\mathcal{I} \setminus \{j \},c), \text{ for every } j \in \mathcal{I}. \label{eqn:min-batch-monotone}
\end{split}
\end{align}

For the induction step, let $\mathcal{B} = (B_1, B_2, \ldots, B_b)$ be a sequence of batches constructed by the algorithm GAC+ for $\mathcal{I}$, $B_1$ being the first batch constructed by the algorithm and $p \in \mathbb{N}$ being the cumulative batch processing time of $\mathcal{B}$.
Let the label of $B_1$ be the job $i = [mint_i, maxt_i]$, i.e., $mint_i$ is maximal among the minimal processing times and the processing time of $B_1$ is equal to $mint_i$.
For the set of jobs $\mathcal{I} \setminus B_1$, the algorithm constructs the batch sequence $B_2, \ldots, B_b$ (see the definition of GAC+).
By the induction hypothesis we know that $B_2, \ldots, B_b$ is optimal for $\mathcal{I} \setminus B_1$, i.e., $b(\mathcal{I} \setminus B_1,c) = \vert \mathcal{B} \vert -1 = b-1$ and $p(\mathcal{I} \setminus B_1,c) = p - mint_i$.
It thus suffices to show that there exists a batch sequence of minimal length and with minimal batch processing time that contains the batch $B_1$.

Let $O_1$ be the batch containing $i$ in an optimal sequence of batches $\mathcal{O}$ and let us choose $\mathcal{O}$ such that the size of the intersection $\vert O_1 \cap B_1 \vert$ is maximal. We will prove that $O_1 = B_1$.

First note that $i \in O_1$ implies that $mint_i \in [mint_j, maxt_j]$ for all jobs $j \in O_1$: $mint_j \leq mint_i$ since $mint_i$ is maximal and $mint_i \leq maxt_j$ since every job $j \in O_1$ needs to be compatible with $i$. Thus the processing time of batch $O_1$ is equal to $mint_i$.

We now distinguish two cases: $\vert B_1 \vert < c$ and $\vert B_1 \vert= c$, where $c$ is the maximum batch size. If $\vert B_1 \vert< c$, the batch $B_1$ contains all neighbors of $i$ in the compatibility graph $G$ corresponding to the set of jobs $\mathcal{I}$. Since $O_1$ is a clique containing $i$, it follows that  $O_1 \subseteq B_1$.
Then by monotonicity (as stated in equation~\eqref{eqn:min-batch-monotone}), we have
\begin{align*}
\vert \mathcal{B} \vert -1 & = b(\mathcal{I} \setminus B_1,c) \leq b(\mathcal{I} \setminus O_1,c) = \vert \mathcal{O} \vert -1 \\
\text{ and } p - mint_1 & = p(\mathcal{I} \setminus B_1,c) \leq p(\mathcal{I} \setminus O_1,c) = p(\mathcal{I},c) -mint_i ,
\end{align*}
which proves that $\mathcal{B}$ is optimal both in terms of the number of batches and in terms of the cumulative processing time.

For the case $\vert B_1 \vert= c$, we assume towards a contradiction that there exists a job $j=[mint_j, maxt_j] \in B_1 \setminus O_1$.
This implies that there must also exist a job $k=[mint_k, maxt_k] \in O_1 \setminus B_1$. (As before, $ O_1 \subset B_1$ would imply that $\mathcal{B}$  is optimal. However, the job $j$ could have been added to $O_1$ without having an impact on the number of batches or the batch processing time required by $\mathcal{O}$. This however is a contradiction to the choice of $\mathcal{O}$).
From the definition of the algorithm, we know that $B_1$ consists of the first $c$ jobs containing $mint_i$. Therefore, $j < k$ and $mint_j \geq mint_k$. Moreover, as noted earlier, we know that $mint_i \in k=[mint_k, maxt_k]$ and thus $[mint_j,mint_i] \subseteq k$.

We then define $O_1' \coloneqq (O_1 \setminus \{ k \}) \cup \{j\}$ and redefine the batch $O \in \mathcal{O}$ that contains $j$ as $O' \coloneqq (O \setminus \{ j \}) \cup \{k\}$.
Both these batches fulfill the compatibility constraint for the processing times: $O_1'$ does because $mint_i$ is contained in $j$ and in all jobs in $O_1$ and $O'$ does because all jobs that are compatible with $j$ are also compatible with $k$ (If job $s$ is compatible with $j$, 
this means that $s=[mint_s, maxt_s] \cap [mint_j,mint_i] \neq \emptyset$, since $mint_i$ is maximal among all minimal processing times. On the other hand, we already noted that 
$[mint_j,mint_i] \subseteq k$ and thus $s \cap k \neq \emptyset,$ which means that $s$ and $k$ are compatible.)
As for the processing times of the batches, both batches $O_1$ and $O_1'$ have the processing time $mint_i$ as they contain job $i$.
For the batch $O'$, we have replaced the job $i$ with a job with smaller or equal processing time ($mint_i \geq mint_k$). Thus the processing time of batch $O'$ is smaller or equal to the batch processing time of $O$.
We have thus produced another optimal sequence of batches $\mathcal{O}'= \mathcal{O} \setminus \{ O_1, O\} \cup \{O_1', O'
\}$.
However, $\vert  O_1'\cap B_1 \vert > \vert  O_1\cap B_1 \vert$ which is in contradiction to the choice of $\mathcal{O}$. This finishes the proof. 
\qed 
\end{proof}

\subsection{Detailed example for the calculation of lower bounds}
\label{sec:appendix-example}

We consider the example instance described in \Cref{sec:example-osp-instance} to exemplify the calculation of the problem-specific lower bounds on the objective function as derived in Section~\ref{sec:lower-bounds}.
    
The values of the lower bounds for the number of batches required and the cumulative batch processing times are summarized in Table~\ref{tab:ex-lower-bounds} on page \pageref{tab:ex-lower-bounds}. We explain their calculation in what follows.
The sets of large jobs are $J_1^l = \emptyset \text{ and } J_2^l = \{ 1, 2, 3, 6\}$, we thus need 4 batches for the large jobs of attribute 2 and none for attribute 1. The processing times for large batches are given by the minimal processing times of the large jobs and contribute $11 + 10 + 19 + 19 = 59$ to the cumulative batch processing time.

For the processing time of small jobs, we exemplify the calculation of the bound based on eligible machines for attribute 1 and the one of the bound based on compatible processing times for attribute 2.
For attribute 1, we have three small jobs (4, 9, and 10) of which job 4 can only be processed on machine 1 and job 9 only on machine 2. Two different batches are thus required for these jobs.
Since the cumulative remaining machine capacity ($2\cdot \max\{c_m\}-(s_4+s_9)=40-(2+4)=34$) is sufficient to accommodate job 10 with $s_{10}=14$, these two batches suffice.
In this case, the runtime of the two batches is given by the minimal runtime of the two jobs 4 and 9, and is equal to 38 in total.
As for attribute 2, the small jobs are 5, 7, and 8. 
Their respective intervals of possible processing times are $[10,50]$, $[11,50]$ and $[50,50]$. 
To follow algorithm GAC+, we sort the list of jobs in decreasing order of their minimal processing times: $(8, 7, 5)$.
A first batch with a processing time of $50$ is created for job $8$. 
The remaining capacity in this batch is $20-11=9$ (assuming that it is assigned to the batch with maximal capacity).
We thus proceed in the list of jobs and add 9 of the 11 units of job 7 to this batch.
For the remaining 2 units of job 7, a new batch with processing time $11$ is created. 
We can add the entire job 5 to this batch.
In total, two batches with a cumulative processing time of $61$ are needed for the small jobs of attribute 2.

For the calculation of setup costs, equation~\eqref{eqn:lower-bound-setup-costs-before-batch} gives:
\begin{align*}
sc &\geq b_1 \cdot \min_{s}\{ sc(s,1) \} + b_2 \cdot \min_{s}\{ sc(s,2) \} \\
& = 2 \cdot \min(6,10) + 6 \cdot \min(8,10)= 60.
\end{align*}
For equation~\eqref{eqn:lower-bound-setup-costs-after-batch}, the list of minimal setup costs \texttt{setup\_costs} contains $\min_{s}\{ sc(1,s) \}=\min(6,8)$ three times (twice for attribute 1 and once for the initial state of machine 1) and $\min_{s}\{ sc(2,s) \}=\min(10,10)$ seven times (six times for attribute 2 and once for the initial state of machine 2). We take the $b=8$ smallest values from this list and thus have:
\[
sc \geq \sum_{i = 1}^8 \texttt{setup\_costs}(i) = 3 \cdot 6 + 5 \cdot 10 = 68.
\]
We take the maximum of these two values and obtain that $sc \geq 68$ for this instance.

Due to the given machine availability intervals for this instance, all jobs except jobs 5, 7, and 8 always finish late. 
Thus, the number of tardy jobs is $\geq 7$ in any feasible solution.

The theoretical lower bound values are reported in Table~\ref{tab:ex-lower-bounds}.

\begin{table}[ht]
    \centering
    \caption{Lower bounds and optimal values for the number of batches, cumulative batch processing time, setup costs, and tardiness for the example instance with 10 jobs.}
    \scriptsize
    \begin{tabular}{l|ccc || ccc || cc || c}
    & \multicolumn{3}{c||}{number of batches} & \multicolumn{3}{c||}{batch processing time} & \multicolumn{2}{c||}{setup costs} & tardiness \\
    & \eqref{eqn:lower-bound-machine-cap} 
    & $b_r^E$\eqref{eqn:even-better-lower-bound-small-jobs-elig-machines}
    & $b_r^C$\eqref{eqn:even-better-lower-bound-small-jobs-proc-times}
     & large jobs & $p_r^E$  & $p_r^C$ \eqref{eqn:even-better-lower-bound-runtime-small-jobs-proc-times} & \eqref{eqn:lower-bound-setup-costs-before-batch} & \eqref{eqn:lower-bound-setup-costs-after-batch} & \\\hline
    attribute 1 & \multirow{2}{*}{6}  &  2 & 1  & 0 & 38 & 19 & \multirow{2}{*}{60} & \multirow{2}{*}{68} & 3 (jobs 4, 9, 10)\\ 
    attribute 2 & & 6 & 6  & 59 & 60 & 61 & & & 4 (jobs 1, 2, 3, 6)\\ \hline
    lower bound  & \multicolumn{3}{c||}{8} & \multicolumn{3}{c||}{158} & \multicolumn{2}{c||}{68} & 7\\ \hline
    optimal values & \multicolumn{3}{c||}{8} & \multicolumn{3}{c||}{158} & \multicolumn{2}{c||}{72} & 8 \\
    gap (in \%) & \multicolumn{3}{c||}{0} & \multicolumn{3}{c||}{0} & \multicolumn{2}{c||}{5.5} & 12.5
    \end{tabular}    
    \label{tab:ex-lower-bounds}
\end{table}

Using the weights and aggregating  the lower bounds for three components of the objective function, we obtain that:
\[
\text{obj} \geq \frac{4 \cdot 158/18 + 68/10 + 100 \cdot 7 }{10 \cdot 105} \approx 0.7066.
\]

Considering the optimal solution for this instance presented in \Cref{sec:example-osp-instance}, the gap between the calculated lower bounds and the optimal solution ($t=8$, $p=158$, and $sc=72$) are thus $0 \%$ for the runtime, $5.5\%$ for the setup costs, and $12.5\%$ for the number of tardy jobs; the gap for the aggregated objective function is $11.7\%$ (due to the high weight given to tardy jobs). 

\subsection{Details concerning the experimental setup}
\label{sec:details-exp-setup}

We consider the theoretical lower bounds as presented in \Cref{sec:lower-bounds}. 
The code is implemented in \verb|C#|.
The experiments are executed on a machine featuring an Intel Core i7-1185G7 processor with 3.00GHz. Each run is executed on a single thread.

We consider the construction heuristic proposed by \citet{lackner-2023-exact} (see \Cref{sec:solution-methods-for-osp}).
The solution method is implemented in \verb|C++|. 
The code is compiled with \verb|Clang++15|.
All experiments are executed on a machine featuring 2x Intel Xeon Platinum 8368 2.4GHz 38C, 8x64GB RDIMM. Each run is executed on a single thread.

We consider the exact methods proposed by \citet{lackner-2023-exact} (see \Cref{sec:solution-methods-for-osp}). 
The ``cpopt'' is implemented with CPLEX Studio 22.11, whereas ``mzn-gurobi'' uses Minizinc 2.8.2 Gurobi 10.0.1. 
All experiments are executed on a machine featuring 2x Intel Xeon CPU E5-2650 v4 (12 cores @ 2.20GHz, no hyperthreading).

We consider the \gls{sa} proposed by \citet{lackner-2022-sa} (see \Cref{sec:solution-methods-for-osp}). 
The \gls{sa} is implemented using \verb|EasyLocal++|, a \verb|C++| framework for \gls{ls} algorithms \cite{digaspero-2003-easylocal}.
The code is compiled with \verb|Clang++15|.
All experiments are executed on a machine featuring 2x Intel Xeon Platinum 8368 2.4GHz 38C, 8x64GB RDIMM. Each algorithm is executed on a single thread.
The algorithm is tuned using \verb|irace| (v.3) \cite{lopex-ibanez-2016-irace}. 
We assign \verb|irace| a total budget of $25,500$ experiments. 
Details on the parameter ranges and their final values are reported in \Cref{tab:sa-parameters}.

\begin{table}[ht]
    \centering
    \caption{Parameter configurations for the \gls{sa} algorithm.}
    \label{tab:sa-parameters}
    \centering
    \begin{tabular}{clcc}
    \toprule
    Param. & Description & Range & Value \\
    \midrule     
    $T_{f}$
    & Final temperature. 
    & $0.001$ -- $0.01$
    & $0.004$\\

    $\alpha$
    & Cooling rate. 
    & $0.985$ -- $0.995$
    & $0.988$\\

    $\rho$
    & Accepted move ratio. 
    & $0.05$ -- $0.7$
    & $0.309$\\

    $p_{\text{\gls{swcb}}}$
    &  Prob. of \gls{swcb} move. 
    &  $0$ -- $1$
    & $0.090$\\
     
    $p_{\text{\gls{ib}}}$
    &  Prob. of \gls{ib} move. 
    &   $0$ -- $1$
    & $0.293$\\
 
    $p_{\text{\gls{mvj}}}$
    &  Prob. of \gls{mvj} move. 
    &   $0$ -- $1$
    & $0.328$\\

    $p_{\text{\gls{mjnb}}}$ 
    & Prob. of \gls{mjnb} move. 
    &   $0$ -- $1$
    & $0.289$\\ 
    \bottomrule
    \end{tabular}
\end{table}

\subsection{Evaluation of the upper bounds provided by the construction heuristic}
\label{sec:upper-bound}

It is important to note that if the construction heuristic successfully schedules all jobs, as is the case for all our benchmark instances, the resulting solution is always feasible, making the obtained solution cost an upper bound on the optimal solution cost.

\begin{figure}[ht]
  \centering
  \includegraphics[width=0.85\textwidth]{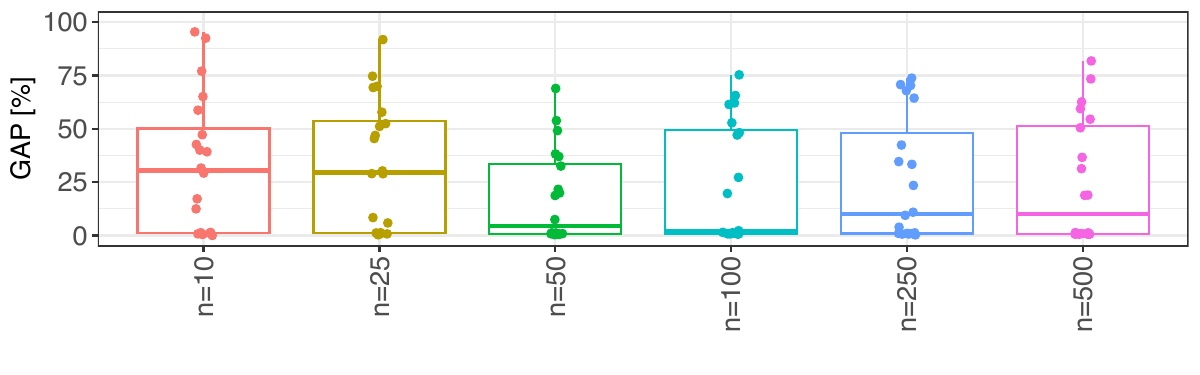}
  \caption{Relative bound gap[\%] between the upper bound found by the construction heuristic and the calculated lower bound per instance considering the overall cost.}
  \label{fig:UC-1-upper-lower-bound}
\end{figure}

We compute the relative bound gap between the calculated lower bound and the cost of the solution generated by the greedy construction heuristic for each benchmark instance (considering the overall cost).
The results are shown in Figure~\ref{fig:UC-1-upper-lower-bound}.
The construction heuristic hardly ever finds optimal solutions (it does so for a single out of the 120 benchmark instances) and often the gap is very large between this upper bound and the calculated lower bound (the relative gap is nearly equal to 100\% for a few instances).
Surprisingly however, for 37 instances across all sizes, the relative bound gap is less than 1\%, even for some of the large instances where no solver could provably find an optimal solution. 
Moreover, for a total of 57 instances, the gap is less than 10\%. 
This suggests that within a short computation time (a maximum of 6 seconds, an average of 0.2 seconds), the construction heuristic together with the problem-specific lower bounds can provide good estimates of the optimal solution cost for a significant portion of the instance set, and rough estimates for nearly half of the instances.
\end{document}